\PassOptionsToPackage{prologue,dvipsnames}{xcolor}
\documentclass[10pt, conference, compsocconf]{IEEEtran}

\usepackage{epsfig}
\usepackage{graphicx}
\usepackage{amsmath}
\usepackage{amssymb}
\usepackage{pgfplots}

\usepackage{url}            
\usepackage{cleveref}
\usepackage{tabularx, booktabs}       
\usepackage{amsfonts}       
\usepackage{nicefrac}       
\usepackage{microtype}      
\usepackage[dvipsnames]{xcolor}
\usepackage[numbers,sort&compress]{natbib}
\usepackage{cleveref}
\usepackage{multirow}
\usepackage{makecell}
\usepackage{soul}

\newcommand{\q}[1]{{\color{black}#1}}

\newcommand{\mypara}[1]{\vspace{2pt}\noindent\textbf{#1}}

\begin{document}
\title{Evaluating 3D Shape Analysis Methods for Robustness to Rotation Invariance}
\author{\IEEEauthorblockN{Supriya Gadi Patil}
\IEEEauthorblockA{Simon Fraser University\\
spandhre@sfu.ca}
\and
\IEEEauthorblockN{Angel X. Chang}
\IEEEauthorblockA{Simon Fraser University\\
angelx@sfu.ca}
\and
\IEEEauthorblockN{Manolis Savva}
\IEEEauthorblockA{Simon Fraser University\\
msavva@sfu.ca}
}
\maketitle

\begin{abstract}
This paper analyzes the robustness of recent 3D shape descriptors to SO(3) rotations, something that is fundamental to shape modeling. Specifically, we formulate the task of rotated 3D object \emph{instance detection}. To do so, we consider a database of 3D indoor scenes, where objects occur in different orientations. 
%
%
We benchmark different methods for feature extraction and classification in the context of this task.
We systematically contrast different choices in a variety of experimental settings investigating the impact on the performance of different rotation distributions, different degrees of partial observations on the object, and the different levels of difficulty of negative pairs.
Our study, \q{on a synthetic dataset of 3D scenes where objects instances occur in different orientations}, reveals that deep learning-based rotation invariant methods are effective for relatively easy settings with easy-to-distinguish pairs. However, their performance decreases significantly when the difference in rotations on the input pair is large, or when the degree of observation of input objects is reduced, or the difficulty level of input pair is increased. Finally, we connect feature encodings designed for rotation-invariant methods to 3D geometry that enable them to acquire the property of rotation invariance.
\end{abstract}

\section{Introduction}

\begin{figure}
\centering
    \includegraphics[width=\linewidth]{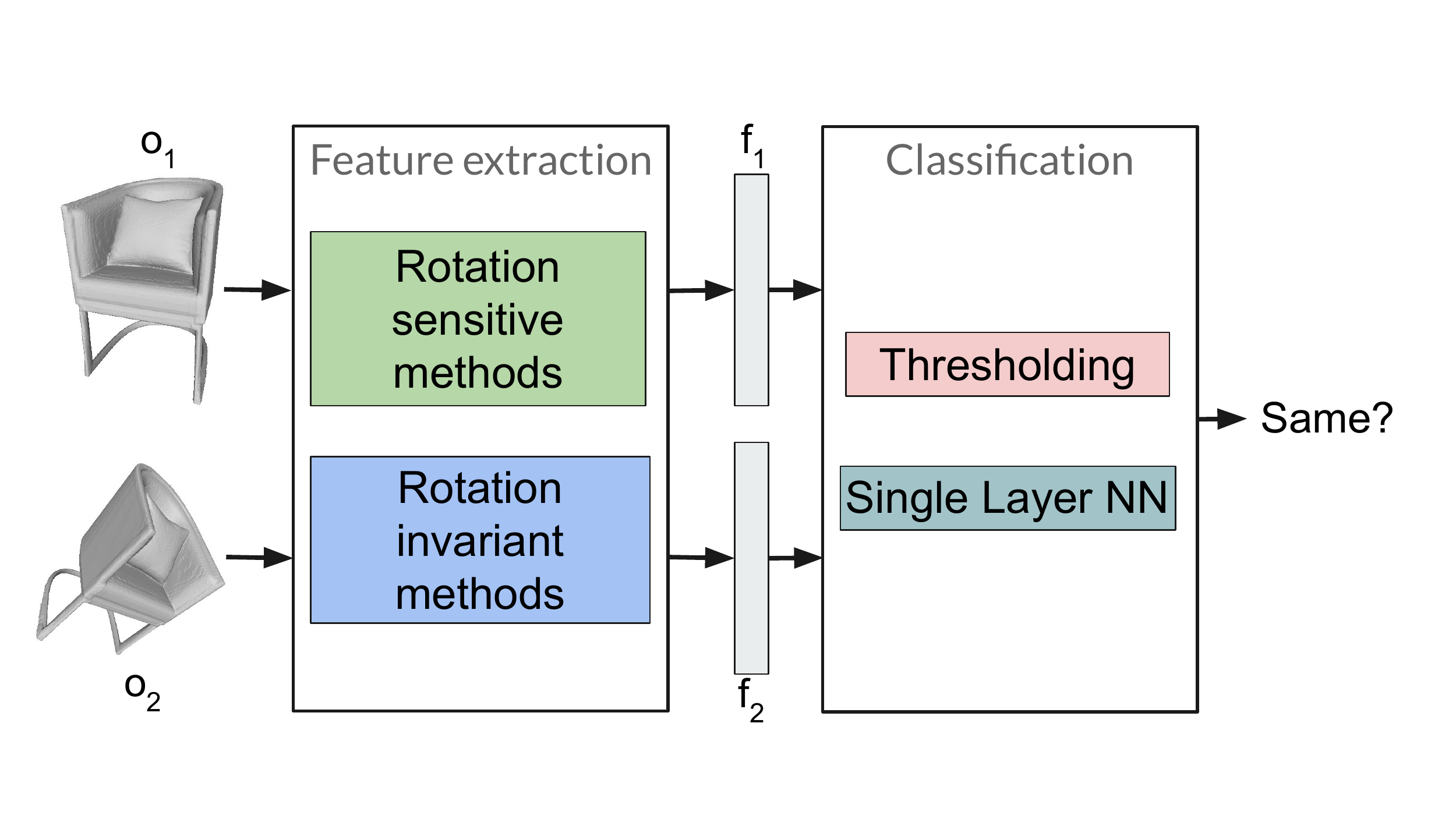}
    \caption{Overview of our analysis setup on the task of instance classification -- a feature extraction module takes a pair of 3D objects and outputs feature vectors (a.k.a shape descriptors; f${_1}$ for o${_1}$, f${_2}$ for o${_2}$).
    A classification module then determines if the given pair is the \emph{same} object or not. 
    In this paper, we evaluate different feature extraction and classification methods (see Section \ref{sec:methodology}) under different \emph{rotation distributions}, different percentages of \emph{partial observation} and different \emph{difficulty levels} of negative pairs  (see Section \ref{sec:experiment_setting} and \ref{sec:results} for details).
    }
    \label{fig:overview}
\end{figure}

Are the two chairs in \Cref{fig:overview} instances of the same object? Answering this question requires an understanding of shape similarity under geometric transformations (rotation operation in \Cref{fig:overview}), which is fundamental to modeling objects in the 3D space. For humans, this is a trivial task, but it is not known to what extent modern deep learning methods for shape analysis are capable of understanding such transformations.
Endowing machines with the ability to decipher these geometric transformations on instances of the same object finds utility in simulation and robotics applications such as grasping, localization, and rearrangement.

A natural question then arises: where are such cases encountered in practice? In the real world, sets of \emph{identical} objects are observed in different orientations; ex., a set of chairs surrounding a dining table or a set of place settings on the table.
In synthetically designed 3D scenes, such sets of objects are typically represented with instances of the \emph{same} 3D model appropriately transformed into various positions.

In this paper, we study the task of rotated object instance detection.
The objective here is to understand the efficacy of traditional \cite{novotni20033d} as well as recent 3D deep learning methods \cite{chen2019learning, mescheder2019occupancy, deng2021vector, chen2022devil, zhao2019rotation, zhang2022riconv++} in the realm of SO(3) rotations on objects.
Having such an understanding can also serve other downstream applications such as sub-scene retrieval, scene editing and scene compression (via indexing and querying), all of which hinge on detecting instances of an object, i.e., 
 whether a pair of 3D object geometries are the \emph{same} object up to a rotation. We formulate this task as a binary classification problem (same or not same) for a pair of input 3D objects.

Though this may initially seem like an easy task, there are a number of challenges.
First, determining whether a rotated pair of object instances are identical up to a rotation can be hard in the presence of geometrically very similar but not identical objects.
Second, objects in 3D scenes span a broad range of categories with wide disparities in the number of instances observed, the distribution over observed rotations, and the geometric variation among instances in the same category.
Lastly, the effect of partial observations of the objects must be taken into account whenever the 3D scene is not complete, such as when the scene is acquired via scanning devices and then reconstructed.

We adopt a pipelined approach where a feature extraction module first computes a feature representation for each input in the pair, and then a classification module outputs the binary label for the feature representation of the pair.
\Cref{fig:overview} illustrates this pipeline and shows the different approaches that we investigate for each module.
%
We benchmark combinations of methods on the task of instance classification in a variety of experimental setups with different rotation distributions, different levels of object partial observation, and different levels of difficulty for an input pair.

Our study shows that modern deep learning-based rotation invariant methods are effective for relatively easy settings with easy-to-distinguish pairs of objects.
However, when the degree of observation of input objects is reduced, and/or the difficulty level of the input pair is increased, their performance decreases significantly.
\section{Related work}

\mypara{Shape representations.}
\q{One of the challenges of working with 3D shapes is their representation -- the notion of standard representation is weak since different kinds of shape representations are used for different use cases.} 
These include meshes, voxel grids \cite{oomes19973d}, octrees \cite{wilhelms1992octrees}, multi-view images \cite{su2015multi}, point clouds \cite{qi2017pointnet, qi2017pointnet++}, geometry images \cite{sinha2016deep}, deformable patches \cite{groueix2018papier}, part-based hierarchies \cite{wang2011symmetry,li2017grass}, and implicit fields \cite{chen2019learning, mescheder2019occupancy, park2019deepsdf, michalkiewicz2019deep}.
In this paper, we analyze and benchmark the performance of different shape encoding methods on \emph{instance matching} of 3D objects under various transformations and settings. The methods we consider mainly make use of meshes, voxel grids, point clouds and implicit fields for shape representation.

\mypara{Shape descriptors.}
A 3D shape descriptor captures the underlying geometric information of the shape in the form of a $k$-dimensional feature vector.
Over the years, many approaches have been proposed for developing robust shape descriptors that could be employed in a variety of shape analysis tasks, such as shape retrieval, classification etc.
%
%
Early works \cite{bromley1993signature, johnson1999using, kortgen20033d, belongie2000shape, manay2004integral, pauly2003multi, rustamov2007laplace, sun2009concise, aubry2011wave} on shape analysis are based on hand-crafted geometric features of the shapes, where the focus is on obtaining a descriptor that is invariant to various types of isometries, both internal and external. 
%
%
However, they are not invariant to object rotations. \q{Kazhdan et al.~\cite{kazhdan2003rotation} focus on developing rotation invariant descriptors using Spherical Harmonics and demonstrate its use for the problem of matching 3D shapes. Another work by Novotni and Klein~\cite{novotni20033d} uses 3D Zernike polynomials to represent the input 3D shape in a manner that is invariant to rotation.}
%
In recent years, learning-based methods for obtaining shape descriptors have become popular, mainly for their ability to efficiently model different shape representations with complex geometries, such as probabilistic models \cite{xie2018learning, shi2020unsupervised} and deep autoencoder/generative models \cite{girdhar2016learning, sinha2016deep, yang2018foldingnet, wu2016learning, achlioptas2018learning, han2019view, park2019deepsdf, mescheder2019occupancy, deng2021vector}. While these deep shape descriptors are richer, they are not guaranteed to be invariant to object rotations in the SO(3) space.  
In this paper, we explore a related direction, specifically for rotated versions of 3D objects. Our focus is on analyzing the performance of recent standalone deep shape descriptors \cite{qi2017pointnet++, chen2019learning, mescheder2019occupancy, deng2021vector, zhao2019rotation, zhang2022riconv++, chen2022devil} on \emph{instance matching} of the objects under various data transformations and compare their performance with a traditional 3D shape descriptor~\cite{novotni20033d}.

\mypara{Rotational in/equivariance in 3D point cloud analysis.} 
We observe a trend of coming up with a new architecture in modeling 3D point clouds under invariance and equivariance constraints~\cite{cohen2018spherical, deng2018ppf, zhang2019rotation, sun2019srinet, zhao2020quaternion, you2021prin,  li2021rotation}. \cite{zhao2019rotation} propose a network with two branches, one for global and the other for local rotation invariant feature learning. These features are then combined by an attention-based fusion layer and used for shape classification or part segmentation. \cite{zhang2022riconv++} introduce a rotation invariant convolution operator on point clouds that considers the relationship between a point of interest and its neighbors, as well as the internal relationship of neighbors, to obtain a rotation-invariant representation of the shape point cloud. This method can also capture the local and global context by changing the neighborhood size considered during training. \cite{chen2022devil} argue that the existing rotation invariant methods cannot fully capture the global information due to a lack of pose information and hence, propose a pose-aware rotation invariant convolution operator that learns dynamic kernel weight for each neighbor based on its relative pose to the center. \cite{wang2022art} takes a different approach to achieve rotation invariance and proposes a method to use adversarial training for point cloud classifiers. This method considers the rotation of the point cloud as an adversarial attack and during training, the network learns to defend it, in turn achieving robustness to rotations.\\
These methods, while having their own merits, limit intuition and understanding of their ability on modeling (in/equi)variance constraints to a single 3D object. Hence, we explore this line of research where we formulate the task of evaluating the ability of different methods to identify a pair of rotated instances of the same objects.\\

\mypara{Selected methods.} We specifically use eight relevant and existing 3D shape analysis methods, of which four are rotation-sensitive methods (PointNet++~\cite{qi2017pointnet++}, IM-NET~\cite{chen2019learning}, Occupancy Net (ONet)~\cite{mescheder2019occupancy}, Vector Neuron Network (VNN)~\cite{deng2021vector}), and the other four are rotation-invariant methods (3D Zernike~\cite{novotni20033d}, LGR-Net~\cite{zhao2019rotation}, RI-Conv++~\cite{zhang2022riconv++}, PaRI-Conv~\cite{chen2022devil}). We analyze the rotation invariance-ness of these 3D shape analysis methods for \emph{instance} classification task. That is, given a pair of rotated 3D objects, we benchmark the performance of respective methods in correctly identifying if the pair belongs to the same object (or not), across different metrics -- accuracy, precision, recall, and F1 scores. To the best of our knowledge, this is the first such effort.

\section{Methodology}
\label{sec:methodology}
Our approach for identifying a pair of object instances being the same (or not) involves two steps -- (a) feature extraction, and  (b) classification.
The feature extraction module (\Cref{subsec:feature_extraction}) explores eight different shape analysis methods under different data transformations and settings (see \Cref{sec:experiment_setting} for details). The classification module (\Cref{subsec:classification_module}), which consumes the shape descriptors (interchangeably referred to as features/feature vectors) from the previous step, explores two ways to output a binary indicator for the intended task. In what follows, we provide details of different approaches we investigate in each module.
%
%
\subsection{Feature extraction module}
\label{subsec:feature_extraction}
We consider two categories of 3D shape analysis methods: rotation-sensitive and rotation-invariant methods. For rotation-sensitive methods, we consider four prominent neural models: PointNet++~\cite{qi2017pointnet++} IMNET  \cite{chen2019learning}, Occupancy Net (ONet) \cite{mescheder2019occupancy} and Vector Neuron Network (VNN) \cite{deng2021vector}. For rotation-invariant methods, we consider three recently proposed neural models: LGR-Net~\cite{zhao2019rotation}, RI-Conv++~\cite{zhang2022riconv++} and PaRI-Conv~\cite{chen2022devil}. We also compare against a traditional 3D shape descriptor, Zernike descriptors~\cite{novotni20033d}. The motivation for selecting these \emph{eight} models for analysis is described below.

\mypara{Rotation-sensitive methods:}

\mypara{(a) PointNet++} ~\cite{qi2017pointnet++} introduce a hierarchical neural network for processing 3D point clouds. Due to the ease of use and availability of scanners, point clouds have become a more common representation for 3D data. As a pioneering work in 3D point cloud analysis, we analyze this model.


\mypara{(b) IMNET.}
\cite{chen2019learning} learns implicit fields for generative shape modeling, realized via an autoencoder-based training setup. We consider this work (and ONet \cite{mescheder2019occupancy} described below) in our analysis since implicit fields can efficiently model the iso-surface of 3D shapes with complex geometries, leading to robust shape descriptors.
We essentially use the same autoencoder training setup as in the original work. Shape descriptors are obtained at the output of the encoder.

\mypara{(c) ONet.}
Similar to IMNET, Occupancy Network (ONet)\cite{mescheder2019occupancy} learns occupancy fields for 3D shape modeling, again realized via an autoencoder-based training setup. We use the architecture proposed for point cloud completion task and consider the output of the encoder as our shape descriptor. We consider the above two works since we want to understand the robustness of these implicit shape descriptors for rotation invariance tasks, such as ours  (i.e., on instance classification of the \emph{same} 3D object under rotational transformations).


\mypara{(d) VNN.}
Vector Neuron Network (VNN) \cite{deng2021vector} proposes a set of novel neural layers that are equivariant to SO(3) rotations, i.e. the rotation applied to input is retained by the output. The overall architecture is similar to that of ONet \cite{mescheder2019occupancy}, except that the traditional scalar neurons are now replaced with what are called as the vector neuron layers. Such rotational equivariance helps in better reconstruction of 3D shapes under rotational transformations in the SO(3) space. Thus, we consider the shape embeddings obtained from VNN as one of the analysis methods.

\begin{figure}[!t]
    \centering
    \includegraphics[height=0.6\columnwidth, width=\columnwidth]{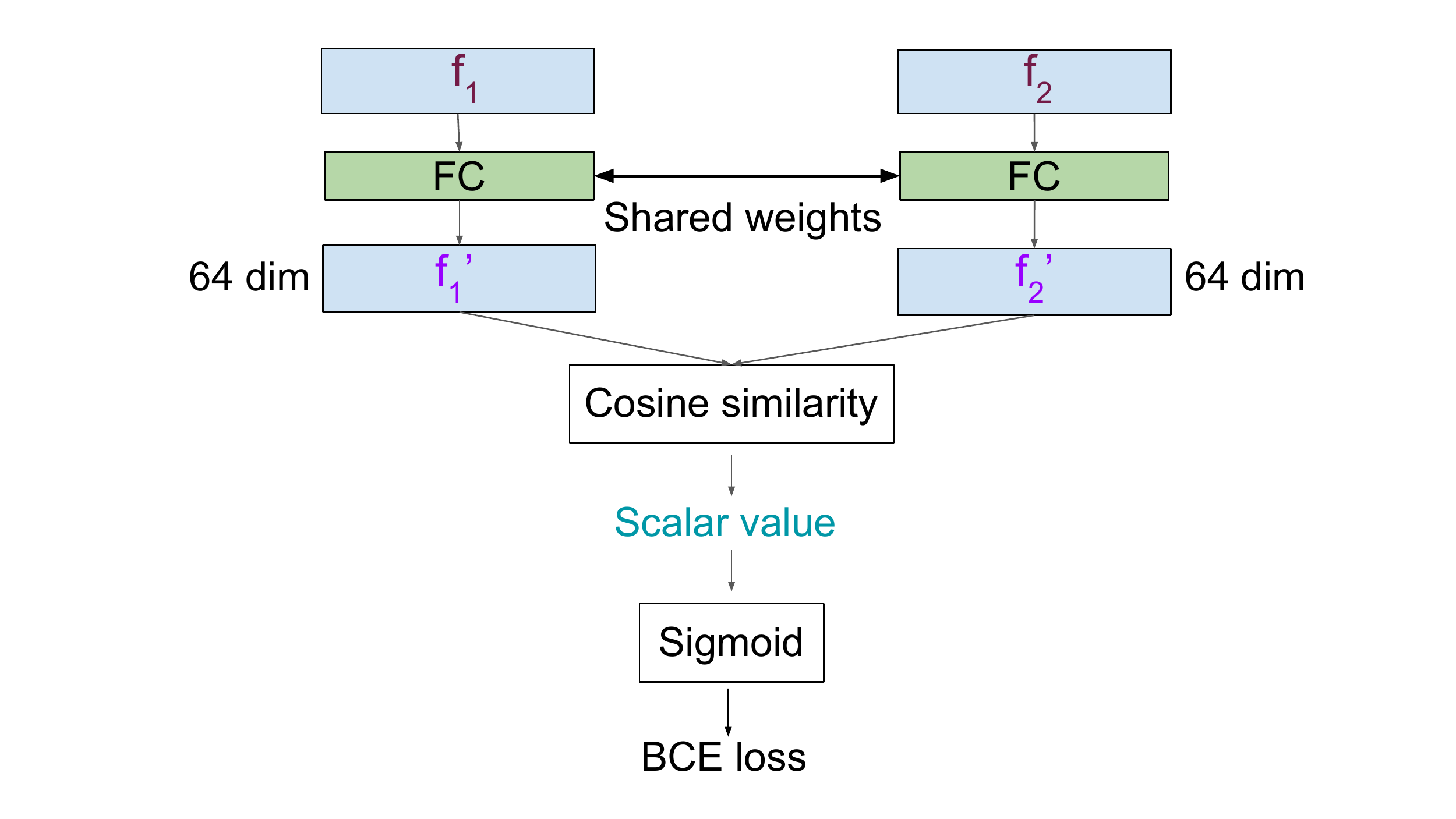}
    \caption{A neural classifier based on a fully-connected layer that consumes a pair of shape descriptors belonging to two 3D shapes and outputs a binary value indicating if they are the same object or not. This is referred to in the paper as 1NN (see Section \ref{subsec:classification_module}).}
    \label{fig:single_NN}
    \vspace{-10 pt}
\end{figure}

\mypara{Rotation-invariant methods:}

\mypara{(a) 3D Zernike descriptors.}
3D Zernike polynomials ~\cite{novotni20033d} provide a way to represent a 3D object that is \emph{invariant} under rotation. 
The coefficients of 3D Zernike polynomial are used as the shape descriptors.
We consider this method as one of the shape encoding methods mainly because it calculates rotation-invariant features in a traditional analytical fashion and does not require any training (unlike other deep learning-based methods), making it a good baseline. 

\mypara{(b) VNN-G.}
It is possible to make VNN features rotation-invariant by computing the Gram matrix ($MM^\top$ for any matrix $M$) of those features. To understand this, let's consider two equivariant signals $V$ and $T$. A signal is said to be equivariant if it is the output of an equivariant function (so, VNN features are a good example of an equivariant signal). From the property of equivariant signals, $VT^\top$ is rotation invariant (refer to \cite{deng2021vector} for details). Thus, the Gram matrix provides rotation-invariant features. We refer to these features as VNN-G in the rest of the paper.

\mypara{(c) LGR-Net.}
Local Global Representation Network (LGR-Net)\cite{zhao2019rotation} proposes a network for learning rotation invariant features on point clouds. The network has two branches, one for global and the other for local rotation-invariant feature learning. The network uses attention-based fusion layer to combine these features. 

\mypara{(d) RI-Conv++.}
Rotation Invariant Convolution (RI-Conv++)\cite{zhang2022riconv++} proposes point cloud-based convolutional network that learns rotation invariant features from local regions. It considers the relationships between a reference point ($r$) and its neighbor ($i$), as well as the relationship among neighbors, to learn informative features that are used in training the network for rotation-invariant task settings.

\mypara{(e) PaRI-Conv.}
Pose-aware Rotation Invariant Convolution (PaRI-Conv)~\cite{chen2022devil} proposes to encode relative pose information between points in the point cloud. The learned features preserve the geometric relationship among neighborhood regions by incorporating pose information.

The last three methods mentioned above are the most recent methods proposed for rotation-invariant feature learning for shape analysis tasks. Hence, we consider them in our analysis.

\subsection{Classification module}
\label{subsec:classification_module}
This module consumes a pair of shape descriptors corresponding to a pair of rotated 3D objects, and outputs a binary indicator of \emph{same-object} or \emph{not-same-object} label. 
A binary output can be obtained in the following two ways.

\mypara{Thresholding.}
\label{sec:thresholding}
A straightforward way to get binary indicators is to threshold the cosine similarity value. But, how to decide on the optimal threshold value?
To this end, we use Youden’s J statistics~\cite{youden1950index} which is defined as J = True Positive rate minus False Positive rate. 
We then consider the threshold with maximum J statistics as the optimal threshold. If the cosine similarity value is above this optimal threshold, we label the pair as the \textit{same-object}, otherwise \textit{not-same-object}. We refer to such a setup as \emph{Thresholding}.

\mypara{Single layer neural network.}
We also consider a learning-based scheme to get a binary output for classification, see Figure \ref{fig:single_NN}. We essentially train a fully connected (FC) layer by employing a sigmoid activation on the cosine similarity value and compare its output with the ground truth. We use the binary cross-entropy loss and backpropagate the gradients. In further sections, we refer to this setting as 1NN.

\section{Dataset}
\label{sec:dataset}
\begin{figure}[t]
    \centering
    \includegraphics[height=0.6\columnwidth, width=\columnwidth]{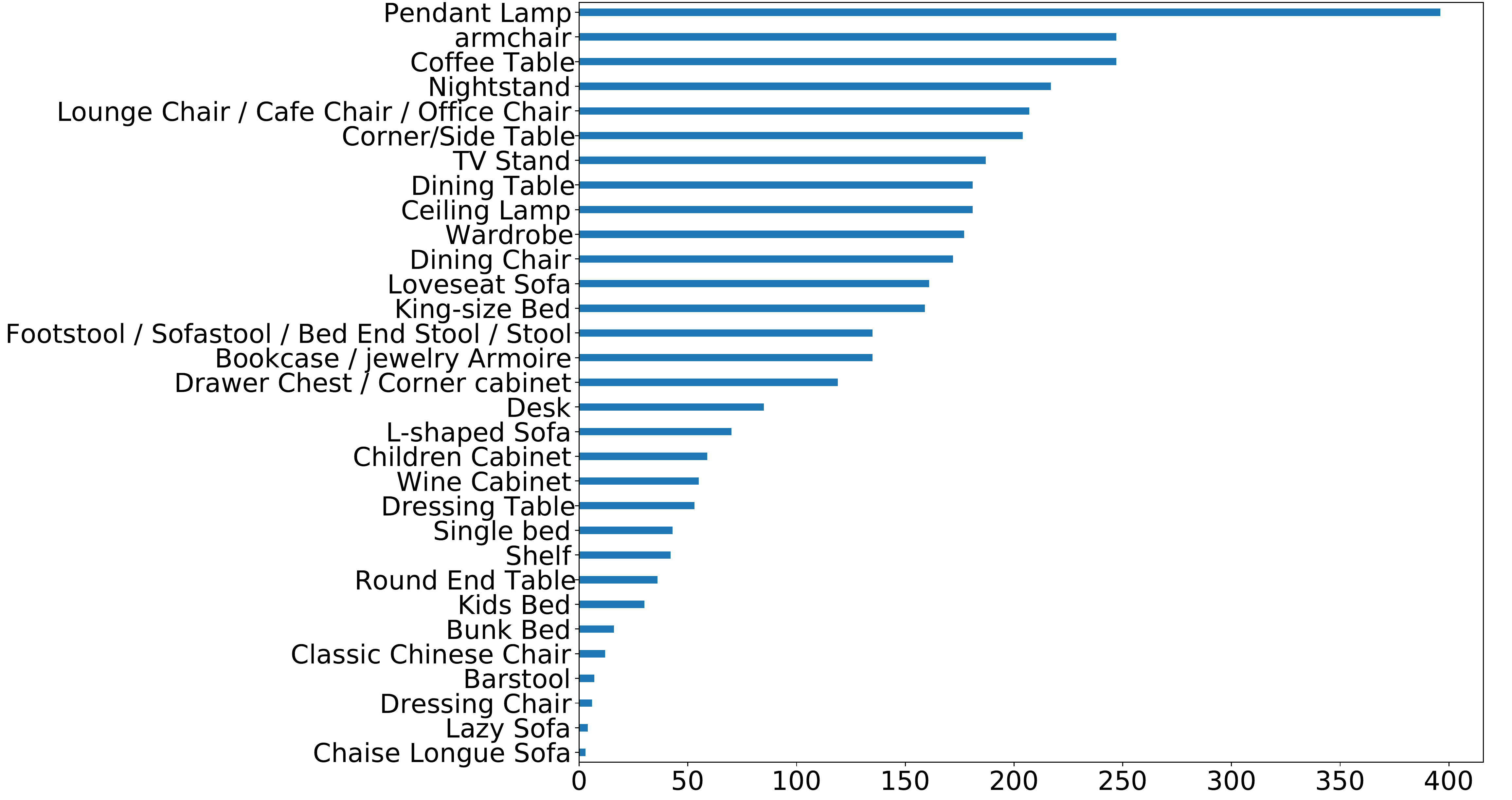}
    \caption{Distribution of unique objects across 34 categories from the 3D-FRONT dataset. Objects in each of these categories repeat across many different scenes, and occur in different orientations, making this dataset well-suited for analyzing different shape analysis methods for their ability in modeling object rotation invariance-ness. 
    }
    \label{fig:freuency_plot}
    \vspace{-10 pt}
\end{figure}
We consider 1500 indoor scenes from the 3D-FRONT dataset \cite{fu20213d}, which contain 3647 unique 3D objects from 34 categories. \Cref{fig:freuency_plot} shows a category-wise distribution of all objects across our database of scenes. We use 80-20 split for train-test sets, respectively, wherever applicable. 
%

For instance classification task, we create positive and negative pairs of objects and the corresponding ground-truth labels. A positive pair is created from objects with the same \emph{instance ID}, whereas a negative pair is created in the following three ways: (i) an \emph{easy negative} pair is created from objects belonging to any two distinct categories, e.g. a table and a sofa, (ii) a \emph{medium negative} pair is created from objects belonging to the same coarse category, but different fine-grained category; e.g. a coffee table and a dining table, and (iii) a \emph{hard negative} pair is created from objects belonging to same fine-grained category e.g. a pair of coffee tables, one with, say, a circular top and other with a square top. 
%

From the train-test split, we respectively create 4K-1K pairs each of positive and negative examples, for a total of 10K such pairs. 
Since there are three different levels of negative examples (i.e., easy/medium/hard), we create such negative pairs for all three difficulty levels.
\section{Experiments}
\label{sec:experiment_setting}
The design of our experiments for instance classification task revolves around three axes -- (a) types of object rotations, (b) degree of object observations (partial/whole, and if partial, how much?), and (c) different levels of difficulty of negative pairs.
Specifically, we consider four different cases of object rotations, four different cases of partial object observations, and three levels of difficulty of negative pairs, yielding forty-eight ($4 \times 4 \times 3$) different experimental settings.  

For object rotations, the four cases include: (i) no rotation, i.e., both 3D shapes in a pair are aligned, (ii) rotation along the up-axis i.e. both shapes are rotated by \emph{different} angles along the up-axis, (iii) SO(3) rotations: both shapes are rotated using different SO(3) rotation matrices, and (iv) rotations from indoor 3D scenes: considering rotation matrices of shapes as they occur in  3D-FRONT scenes \cite{fu20213d}.

\begin{figure}[t]
    \centering
    \includegraphics[height=0.6\columnwidth, width=\columnwidth]{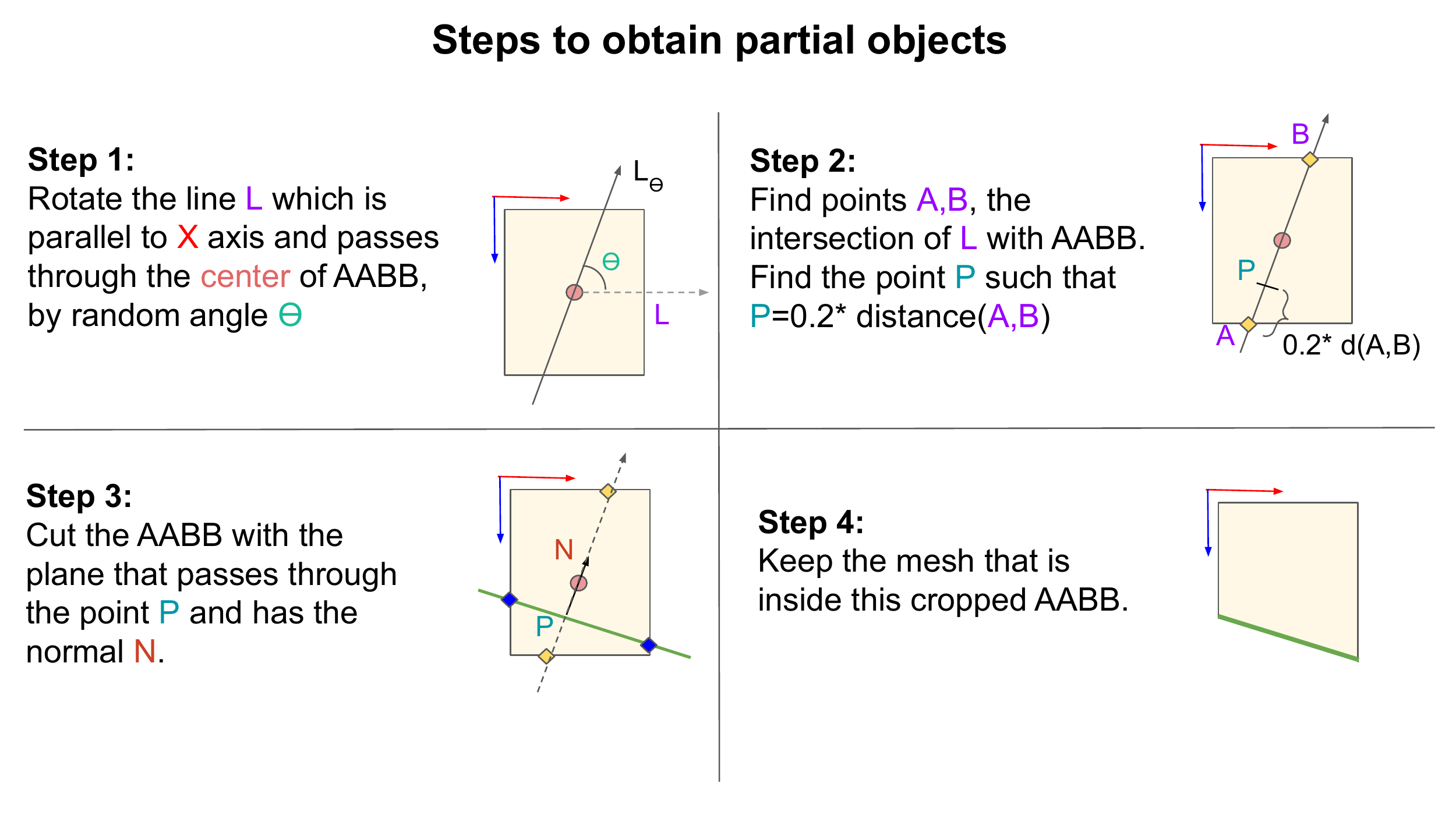}
    \caption{\emph{Top-view} visualizations showcasing different steps to obtain partial observations of 3D objects. In the above figure, we cut 20\% of the Axis-Aligned Bounding Box (AABB), keeping the rest 80\%.} 
    \label{fig:crop_mesh}
    \vspace{-10 pt}
\end{figure}

For partial observations of objects, we consider four cases based on the extent to which the 3D object is cut: (i) 0\% cut i.e. complete objects are considered, (ii) 20\% cut i.e. 20\% of the object is cut and the remaining 80\% is used for our experiments, (iii) 50\% cut and (iv) 70\% cut (remaining 30\% used for our experiments). \Cref{fig:crop_mesh} illustrates such data processing procedure. 

\subsection{Feature extraction module}
The seven deep learning-based 3D shape analysis methods considered in our study are originally trained on the ShapeNet~\cite{chang2015shapenet} or ModelNet40~\cite{wu20153d} object dataset \cite{chang2015shapenet}, which is significantly different from the 3D-FRONT scene dataset used in our experiments. As such, we train all methods, from scratch, on 3D objects from our scene database. We use $28,832$ objects from $1500$ scenes of 3D-Front~\cite{fu20213d} dataset for training, as these objects appear in different scenes under different orientations, and are used as inputs for our analysis task. For context, we briefly touch upon these methods below.

\mypara{PointNet++~\cite{qi2017pointnet++}}. The network is trained on 3D shape classification task. The input is a point cloud with normals. The output from the penultimate layer, which is $256$ dimensional, is used as a feature vector.

\mypara{IMNET~\cite{chen2019learning}.}
We use the original training setup and prepare data as provided on the code repository. 
The trained model produces a $256$-D feature vector for each input 3D object.

\mypara{ONet~\cite{mescheder2019occupancy}.}
We use the pre-processing steps and the model proposed for point cloud completion task as provided in the code repository.
We extract a $513$-D feature vector for each 3D object from the trained model.

\mypara{VNN~\cite{deng2021vector}.}
Vector Neuron network (VNN) learns rotation equivariant features in the SO(3) space. Its network architecture is based on ONet ~\cite{mescheder2019occupancy}, with the difference being the dimensions of network layer. To be specific, if ONet has a layer of size $N$, then, the corresponding layer in VNN has a size equal to $\lfloor \frac{N}{3} \rfloor \times 3$. We use the same pre-processed data from 3D-FRONT that was created as a pat of ONet's data processing step. 
We train the VNN model for point cloud completion task and use same hyperparameters for training as those employed for ONet training. The network obtains a $173\times3$ dimensional features for each object. Since the output is a matrix, we perform a simple `mean' operation to get a vector of $173$-dimension. This is done to keep the comparisons fair with the above feature extraction methods that output vectors.

\mypara{VNN-G.}
We obtain rotation invariant features from equivariant ones by computing the Gram matrix (see \cite{deng2021vector} for details). Specifically, we first extract the $173 \times 3$ equivariant features from VNN and multiplying this feature matrix with its transpose. We do this for every object input to this module. We then take mean across last dimension to get a final feature vector of size $173$ dimensions.

\mypara{LGR-Net~\cite{zhao2019rotation}} The network is trained on 3D shape classification task and takes points and their normals as inputs. For a given point in the point cloud, the \emph{local} branch of the network finds k-nearest neighbors to get a local patch of 32 points. The \emph{global} branch uses farthest point sampling to obtain 32 global points and applies SVD on a downsampled version of these points. The network is trained for 300 epochs with a batch size of 24. The feature vectors of size $256$-D are obtained from penultimate layer of LGR-Net.

\mypara{RI-Conv++~\cite{zhang2022riconv++}}
This network is also trained on 3D shape classification task and uses both point coordinates and their normals as input. It uses farthest-point sampling to sample 32 points. For every point, 32 local neighborhood points are obtained using a k-NN search. From these points, specially designed features are obtained that enable injecting the property of rotation invariance. Here, we consider $128$-D output from the penultimate layer as the feature vector.

\mypara{PaRI-Conv~\cite{chen2022devil}}
This network also takes point clouds with normals as input and is trained for shape classification task. The output from the penultimate layer, which is of $256$-D, is used as a feature vector in our analysis.

\begin{table}
    \centering
    \begin{tabular}{l|c|c}
        \toprule
        \textbf{Methods}  & \textbf{Input format} & \textbf{Originally \emph{trained} for}\\
        \midrule
        \multicolumn{3}{l}{Rotation-sensitive methods}\\
        \midrule
        PointNet++~\cite{qi2017pointnet++} & pc+n & Classification\\
        IMNET~\cite{chen2019learning} & voxel & Reconstruction\\
        ONet~\cite{mescheder2019occupancy} & pc & Reconstruction\\
        VNN~\cite{deng2021vector} & pc & Reconstruction\\
        \midrule
        \multicolumn{3}{l}{Rotation-invariant methods}\\
        \midrule
        3D Zernike~\cite{novotni20033d} & voxel & No training\\
        VNN-G & pc & Reconstruction\\
        LGR-Net~\cite{zhao2019rotation} & pc+n & Classification\\
        RI-Conv++~\cite{zhang2022riconv++} & pc+n & Classification\\
        PaRI-Conv~\cite{chen2022devil} & pc+n & Classification\\
        \bottomrule
    \end{tabular}
    \caption{Table enumerating different input formats used by different 3D shape analysis methods chosen in our analysis. \emph{pc} refers to pointcloud, and \emph{n} refers to their normals. The last column lists different tasks these methods are originally \emph{trained} for.}
    \label{fig:method_metadata}
    \vspace{-10 pt}
\end{table}
\Cref{fig:method_metadata} summarizes different methods in terms of the input and the task they originally address. 
We train these networks on their corresponding tasks using objects from 3D-Front scenes and then consider their features for our analysis.
Despite the fact that these methods use different types of input formats and are trained on different tasks, comparing their performance in our study is fair since they all essentially learn a global geometric representation for 3D shapes, individually tailored with task-specific improvements. 
In our study, we want to analyze these learned geometric representations for robustness to rotation invariance. Plus, these representations are tested on a \emph{common} task of instance classification to measure the rotation invariance-ness, making comparisons fair.

\subsection{Classification module}
Extracted features are now fed to the classification module, which performs a binary classification.
For \emph{Thresholding}, we compute cosine similarity between feature vectors and use Youden's J statistics to find optimal threshold for classification.
When training the 1NN classifier, we use Adam optimizer with a \textbf{lr} of 1e-3 and train for 200 epochs.

\section{Results and Evaluation}
\label{sec:results}

\begin{figure}[t]
    \centering
    \includegraphics[height=0.6\columnwidth, width=\columnwidth]{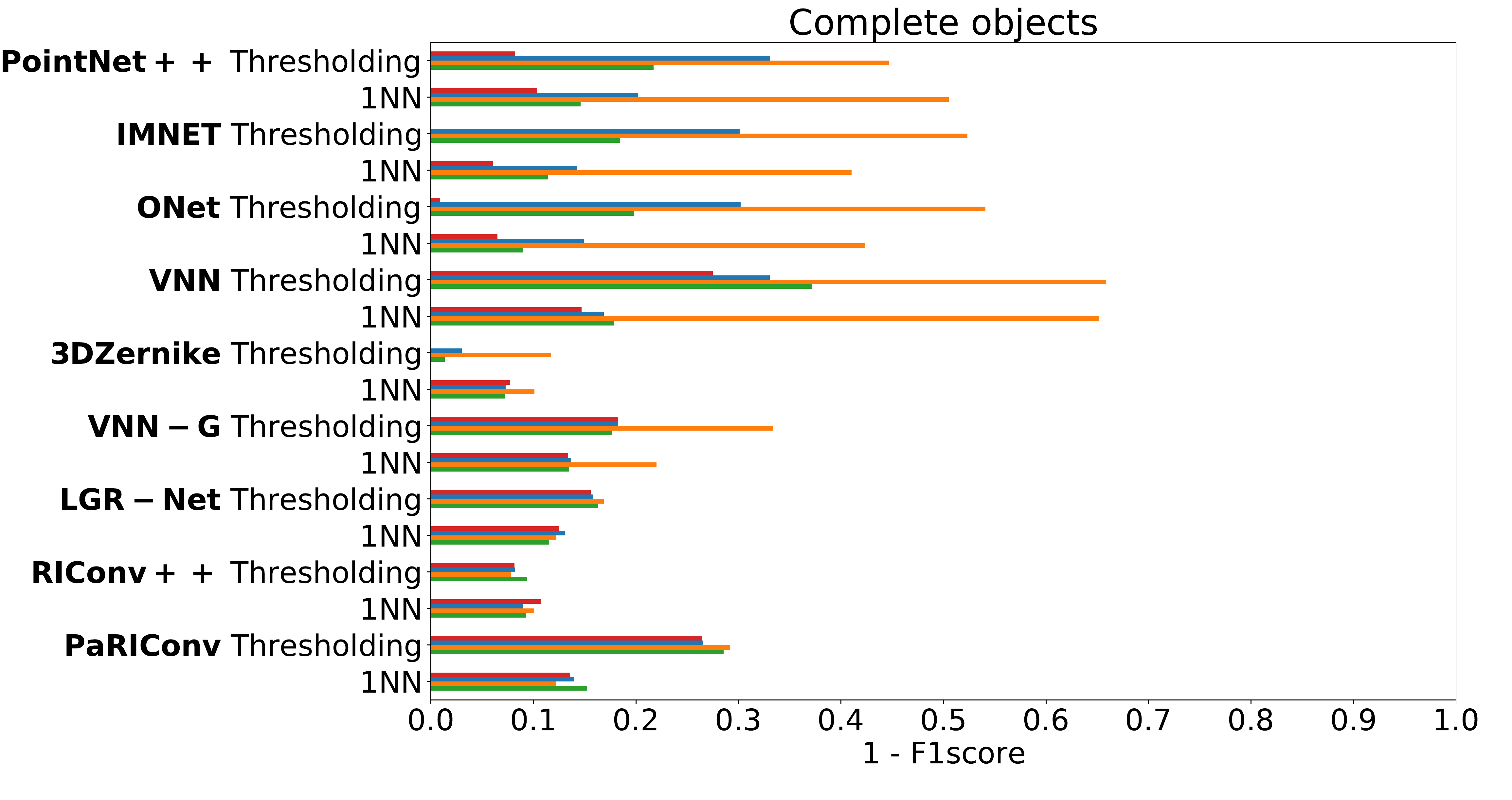}
    \includegraphics[height=0.6\columnwidth, width=\columnwidth]{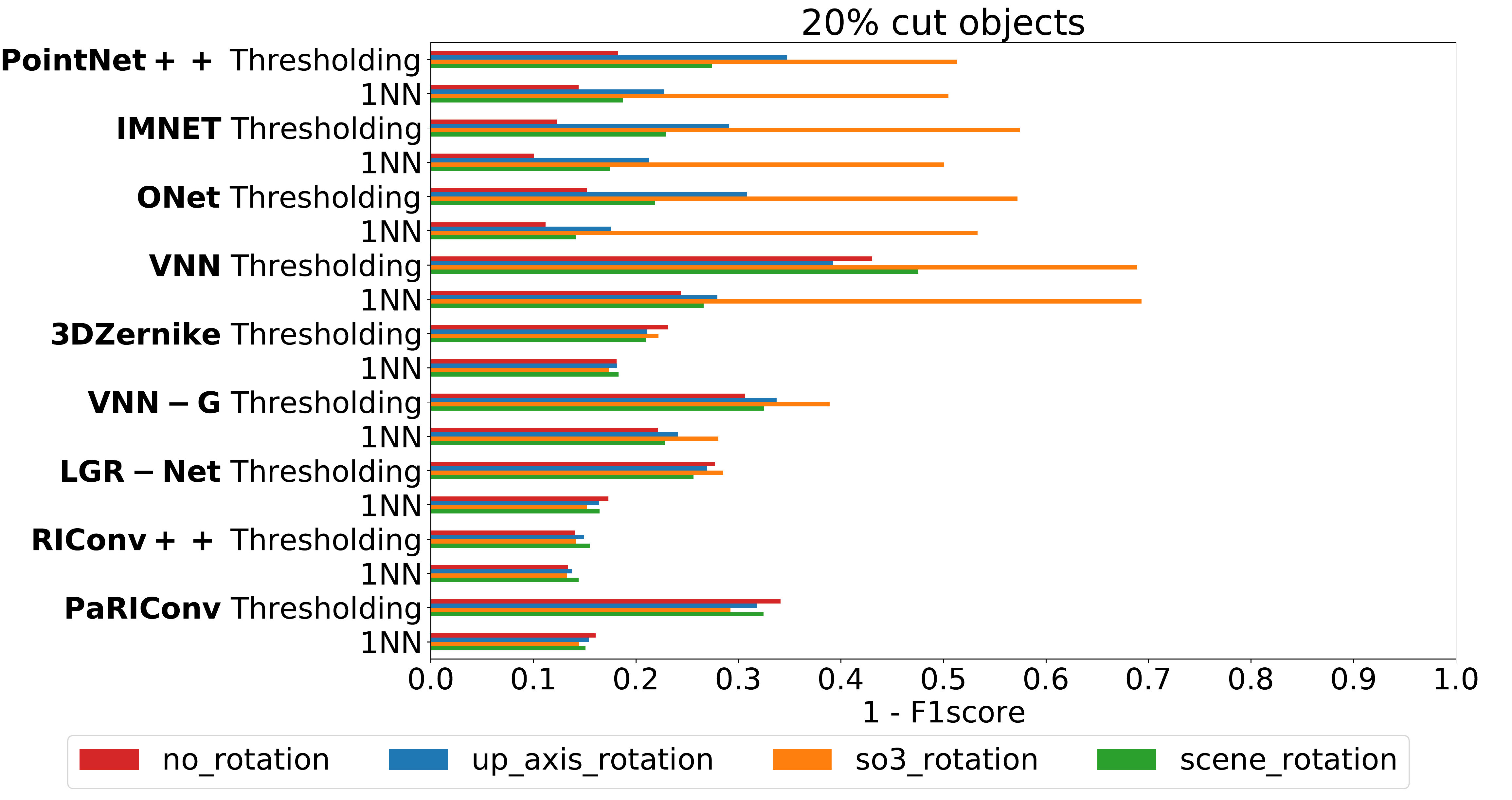}
    \caption{Bar plot shows the error rate of classifying features of hard-negative pairs when
    complete objects and 20\% cropped
    objects are (1) \textcolor{Red}{not rotated}, (2) rotated along \textcolor{NavyBlue}{up-axis}, (3) rotated by \textcolor{Orange}{SO(3)} matrix and (4) rotated according to \textcolor{Green}{scene rotation} matrix. 1NN: training single layer neural classifier.}
    \label{fig:error_bar_plot_complete}
\end{figure}
\begin{figure}[t]
    \centering
    \includegraphics[height=0.6\columnwidth, width=\columnwidth]{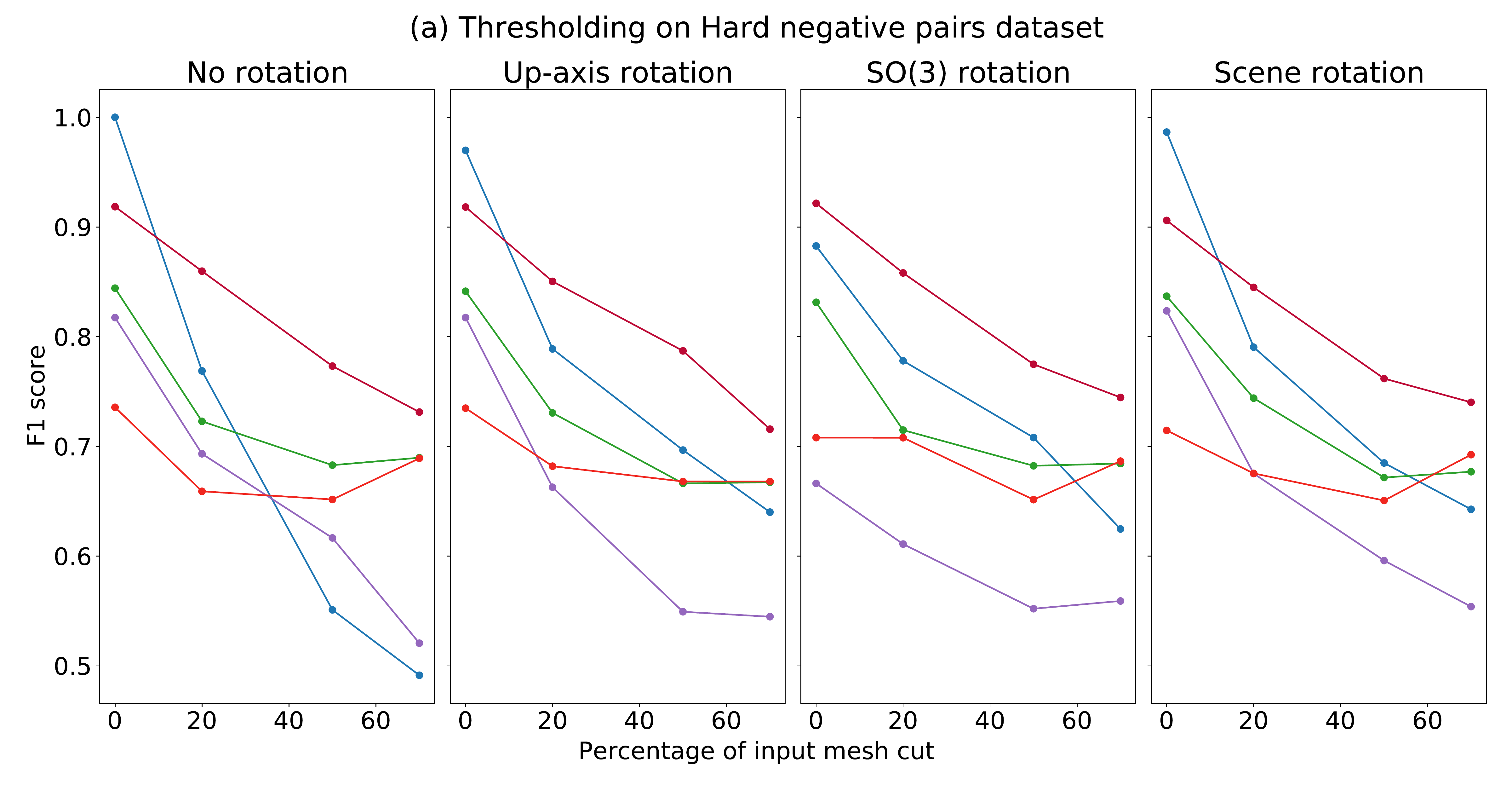}


    \centering
    \includegraphics[height=0.6\columnwidth, width=\columnwidth]{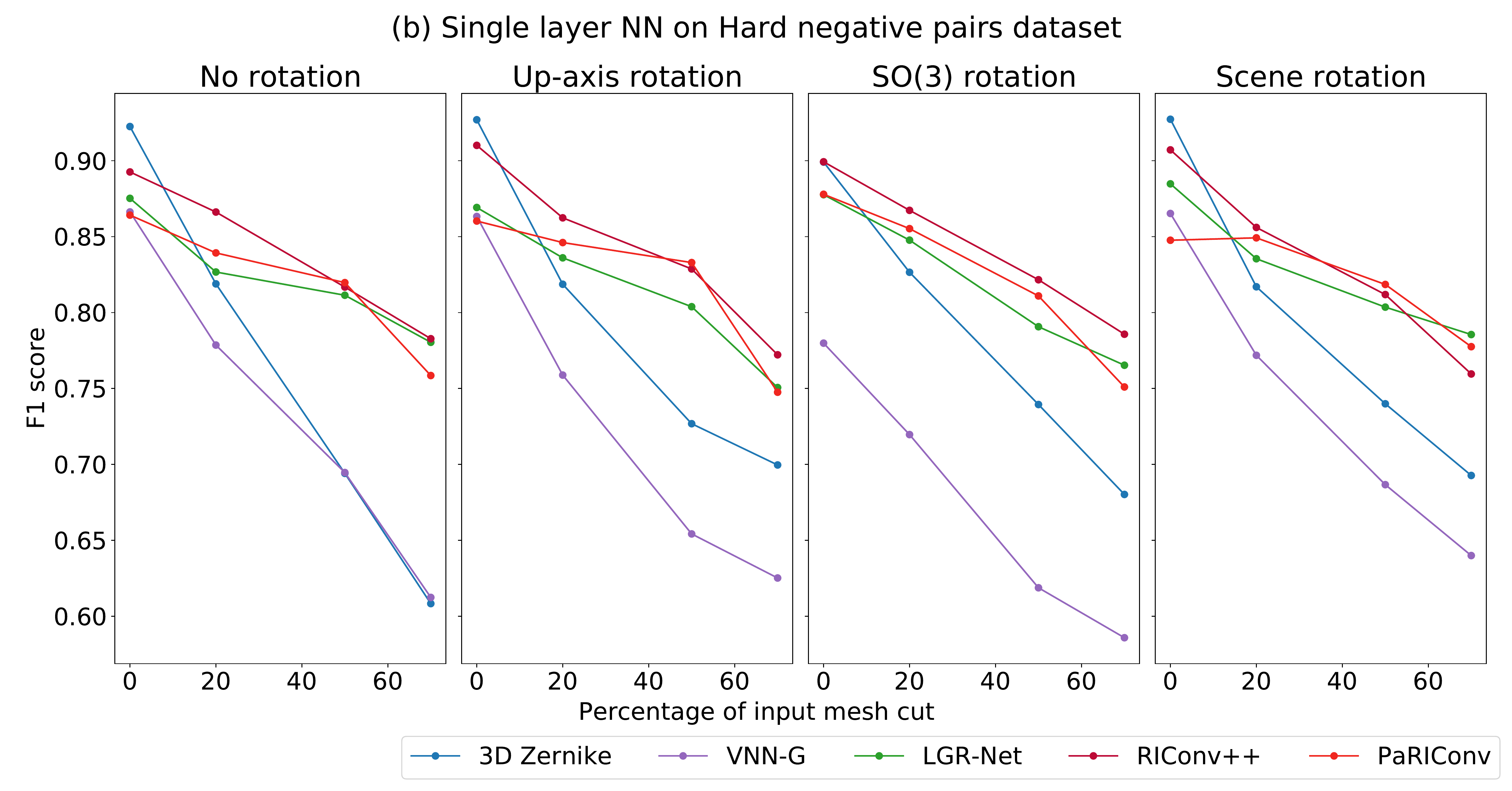}
    \caption{Line plot shows the effect of \emph{hard negative} pairs with different levels of partial input on the classification performance of rotation invariance. The line plots of Figure (a) and (b) respectively show performance of using Thresholding and 1NN classification methods.} 
    \label{fig:line_plot_th_1NN}
\end{figure}

\begin{figure}[t]
    \centering
    \includegraphics[height=0.55\columnwidth, width=\columnwidth]{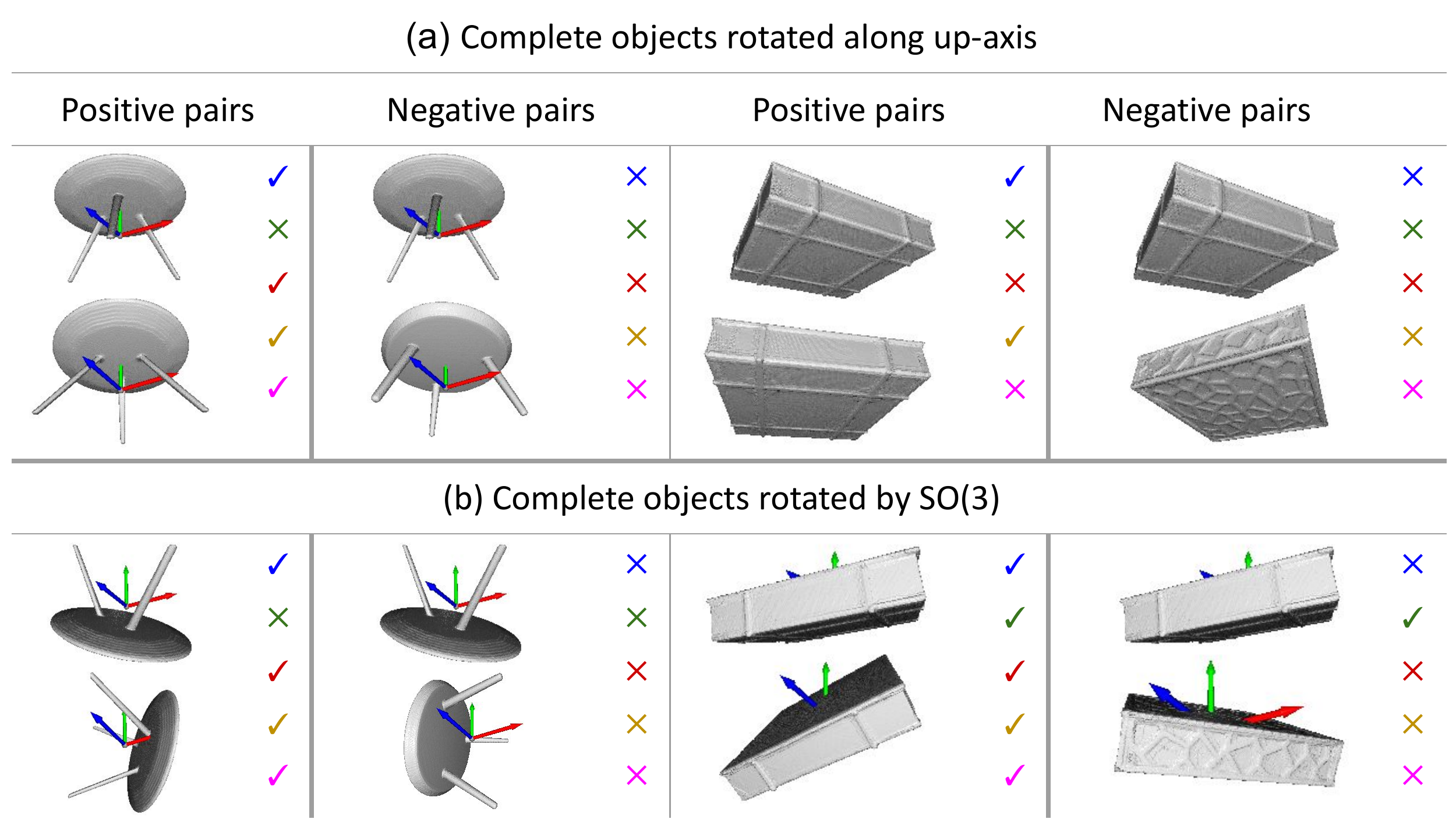}
    \includegraphics[height=0.55\columnwidth, width=\columnwidth]{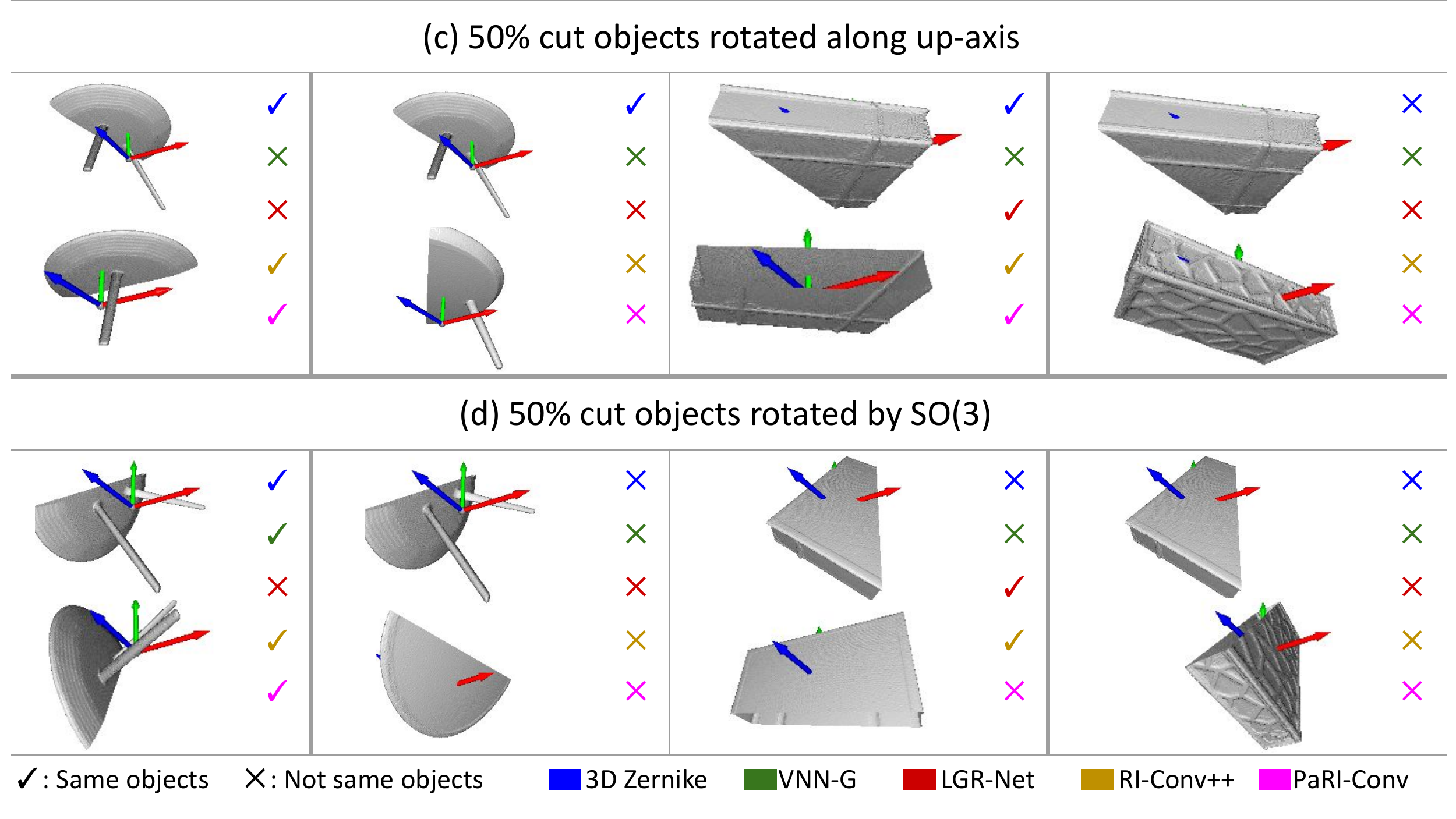}
    \caption{Qualitative results on positive and very hard negative pairs, i.e., distinct objects from the same fine-grained category. The table shows predictions by \textcolor{blue}{3D Zernike}, \textcolor{OliveGreen}{VNN-G}, \textcolor{red}{LGR-Net}, \textcolor{orange}{RI-Conv++}, \textcolor{magenta}{PaRI-Conv}. The ground truth label for positive pairs is \emph{same-object} (denoted by \checkmark) and for negative pairs it is \emph{not-same-object} (denoted by $\times$). Ideally, we want all methods to predict \checkmark for all positive pairs and $\times$ for all negative pairs.}
    \label{fig:qualitative_results}
    \vspace{-13 pt}
    
\end{figure}

We provide extensive quantitative results (Figure \ref{fig:error_bar_plot_complete}, \ref{fig:line_plot_th_1NN} and Table \ref{table:final_half_so3 rotation}) measuring the performance of different 3D shape analysis methods under different types of object rotations, degrees of observations and difficulty levels of negative pairs, on instance classification task. Here, we follow standard evaluation protocols of Accuracy (Acc), Precision (Pre), Recall (Rec) and F1 score \cite{manning2008evaluation}. 
Also, qualitative (Figure \ref{fig:qualitative_results}) and quantitative results (Table \ref{table:final_half_so3 rotation}) over many distinct positive and negative pairs have been presented. 
We want to point out that many different combinations of experimental settings exist (48= 4 types of rotations * 4 types of partial observations * 3 levels of difficulty of negative pairs). However, we only include a sampler of experiments that uncover major settings in a progressive manner. 

First, we consider all types of rotations on 0\% and 20\% cut objects, on all eight 3D shape analysis methods with different classifier settings. This is presented in Figure \ref{fig:error_bar_plot_complete}. 
Second, from the observations of previous experiment, we now consider just SO(3) rotations and increase the partial observation to 50\% of the objects. We analyse all eight 3D shape analysis methods over different positive pairs and different difficulty levels of negative pairs, on different classifier settings. This is tabulated in Table \ref{table:final_half_so3 rotation}. 
And finally, we consider the hard negative pairs alone, and record F1 score by varying the percentage of object-cut applied to the input object from 0 to 70\%, on all types of rotations, using the two classification schemes; see Figure \ref{fig:line_plot_th_1NN}. 

In the subsequent parts, we make systematic observations and provide insights into the observed results when different data transformations and settings are employed.

\begin{table*}[h]
\centering
\begin{tabular*}{0.8\linewidth}{@{}llccccc|ccccc|cccc@{}}
\cmidrule{1-16}
&Difficulty level $\rightarrow$ & \multicolumn{4}{c}{\textbf{Easy}} & & \multicolumn{4}{c}{\textbf{Medium}} & & \multicolumn{4}{c}{\textbf{Hard}}\\
\cmidrule{3-16} & \makecell{Shape Analysis \\methods $\downarrow$} & Acc & Pre & Rec & F1 & & Acc & Pre & Rec & F1 & & Acc & Pre & Rec & F1 \\
\cmidrule{1-16}
\multicolumn{16}{l}{\textbf{Thresholding}}\\
\cmidrule{1-16}
\multirow{4}{*}{\rotatebox{90}{\makecell{Rotation\\ sensitive}}}&PointNet++~\cite{qi2017pointnet++} & 53.3 & 53.1 & 57.9 & 55.4 && 53.2 & 54.2 & 41.9 & 47.3 && 51.9 & 53.3 & 31.1 & 39.3 \\ 
&IMNET~\cite{chen2019learning} & 56.0 & 59.4 & 37.8 & 46.2 && 53.8 & 54.1 & 49.7 & 51.8 && 53.0 & 62.1 & 15.4 & 24.7 \\ 
&ONet~\cite{mescheder2019occupancy} & 54.9 & 57.7 & 36.7 & 44.9 && 51.4 & 51.4 & 55.0 & 53.1 && 50.6 & 50.8 & 43.4 & 46.8 \\ 
&VNN~\cite{deng2021vector} & 55.5 & 70.2 & 19.3 & 30.3 && 54.7 & 67.9 & 17.8 & 28.2 && 54.2 & 63.0 & 20.3 & 30.7 \\ 
\cmidrule{1-16}
\multirow{5}{*}{\rotatebox{90}{\makecell{Rotation \\invariant}}}&3D Zernike~\cite{novotni20033d} & 76.8 & 82.4 & 68.0 & 74.5 && 72.5 & 77.5 & 63.6 & 69.9 && 72.8 & 76.2 & 66.1 & 70.8 \\ 
&VNN-G~\cite{deng2021vector} & 66.4 & 76.5 & 47.3 & 58.5 && 66.0 & 75.5 & 47.4 & 58.2 && 62.6 & 68.8 & 46.1 & 55.2 \\ 
&LGR-Net~\cite{zhao2019rotation} & 85.0 & 82.5 & 88.9 & 85.6 && 71.9 & 75.7 & 64.5 & 69.7 && 69.7 & 71.6 & 65.2 & 68.2 \\ 
&RIConv++~\cite{zhang2022riconv++} & \textbf{88.7} & \textbf{90.4} & 86.6 & \textbf{88.5} && \textbf{81.0} & \textbf{86.5} & \textbf{73.5} & \textbf{79.5} && \textbf{78.8} & \textbf{82.8} & \textbf{72.8} & \textbf{77.5} \\ 
&PaRIConv~\cite{chen2022devil} & 88.0 & 85.0 & \textbf{92.3} & \textbf{88.5} && 68.6 & 71.6 & 61.7 & 66.3 && 67.0 & 69.1 & 61.6 & 65.2 \\ 
\cmidrule{1-16}
\multicolumn{16}{l}{\textbf{Single Layer Neural Network}}\\
\cmidrule{1-16}
\multirow{4}{*}{\rotatebox{90}{\makecell{Rotation\\ sensitive}}}&PointNet++~\cite{qi2017pointnet++} & 66.4 & 70.8 & 55.8 & 62.4 && 58.1 & 63.8 & 37.2 & 47.0 && 56.5 & 61.8 & 33.8 & 43.7 \\ 
&IMNET~\cite{chen2019learning} & 66.6 & 68.1 & 62.4 & 65.1 && 60.6 & 65.7 & 44.3 & 52.9 && 53.6 & 57.3 & 28.6 & 38.2 \\ 
&ONet~\cite{mescheder2019occupancy} & 62.2 & 68.2 & 45.8 & 54.8 && 60.3 & 63.3 & 49.1 & 55.3 && 55.9 & 62.7 & 28.8 & 39.5 \\ 
&VNN~\cite{deng2021vector} & 54.0 & 66.7 & 16.0 & 25.8 && 53.3 & 65.1 & 14.2 & 23.3 && 54.9 & 67.0 & 19.5 & 30.2 \\ 
\cmidrule{1-16}
\multirow{5}{*}{\rotatebox{90}{\makecell{Rotation \\invariant}}}&3D Zernike~\cite{novotni20033d} & 80.2 & 78.4 & 83.6 & 80.9 && 76.9 & 76.5 & 77.6 & 77.1 && 73.8 & 73.4 & 74.5 & 73.9 \\ 
&VNN-G~\cite{deng2021vector} & 75.9 & 80.0 & 69.0 & 74.1 && 72.4 & 78.6 & 61.4 & 69.0 && 68.2 & 77.0 & 51.7 & 61.9 \\ 
&LGR-Net~\cite{zhao2019rotation} & 85.2 & 78.8 & 96.3 & 86.7 && 82.7 & 84.0 & 80.7 & 82.3 && 78.8 & 78.3 & 79.9 & 79.1 \\ 
&RIConv++~\cite{zhang2022riconv++} & \textbf{90.5} & \textbf{85.8} & \textbf{97.0} & \textbf{91.1} && \textbf{85.5} & \textbf{85.5} & \textbf{85.4} & \textbf{85.4} && \textbf{81.8} & \textbf{80.8} & 83.6 & \textbf{82.2} \\ 
&PaRIConv~\cite{chen2022devil} & 87.3 & 81.8 & 96.0 & 88.3 && 82.0 & 80.6 & 84.1 & 82.3 && 80.0 & 76.9 & \textbf{85.8} & 81.1 \\ 
\cmidrule{1-16}
\end{tabular*}
\caption{Test-time classification performance, measured across four different metrics, of different shape encoding methods on input pairs of 3D objects that are 50\% cut, rotated randomly by SO(3) rotation matrix, with three difficulty levels of negative pairs: easy (objects from \emph{distinct} classes), medium (distinct objects from the \emph{same coarse} category), hard (distinct objects from the \emph{same fine-grained} category).}
\label{table:final_half_so3 rotation}
\vspace{-10 pt}
\end{table*}

\mypara{Effect of rotation type.}
\Cref{fig:error_bar_plot_complete} shows the bar plot of error rate, which is 1-F1\_score (1 minus F1 score), for classifying pairs of 0\% and 20\% cut 3D objects. 
Looking at the bar plots, we see that rotations in the SO(3) space (orange bars) incur higher error rates for rotation-sensitive methods (PointNet++, IMNET, ONet, VNN) as compared to rotation-invariant methods (3D Zernike, VNN-G, LGR-Net, RI-Conv++, PaRI-Conv). This indicates that rotation-invariant methods are robust in correctly classifying the input pair of objects for SO(3) rotations. This is an important finding that shows that even pioneering 3D shape analysis methods such as PointNet++, IMNET and ONet are inept in understanding the most fundamental 3D operation, i.e., 3D rotation.

When we consider experiments with up-axis rotation (blue bars) and scene rotations (green bars), we observe that the error rates in these two settings are comparable. This is because objects in indoor scenes are often rotated around the up-axis, e.g. chairs around a dining table.

At last, we observe that the error bars of rotation-invariant methods for all four types of rotations are comparable, but do vary with changes in the rotation axes. This indicates that though the features learned by these methods are rotation invariant, one will outperform others under these settings. This can be narrowed down with controlled perturbations in the input, which we investigate next.

\mypara{Effect of partial input.}
Classification performance via F1 scores when input 3D objects are cut by different percentages is shown in \Cref{fig:line_plot_th_1NN}. Not surprisingly, we observe that with an increase in the percentage of cut applied on the object (i.e., observed surface area is reduced), the performance degrades. Specifically, we see a sharp dip in the performance of the traditional 3D Zernike descriptors. Performances of other rotation-invariant methods are also quite affected. This points to the fact that these deep learning-based methods are sensitive to the relative observations of 3D input. After training a single-layer neural network on top of the feature vectors, we notice an increase in the performance of all methods as compared to directly using features from pre-trained models.
Overall, for any percentage of object cut, RI-Conv++ outperforms all other methods, even in the extreme case of 70\% cut objects.

\begin{figure}[t]
    \centering
    \includegraphics[height=0.55\columnwidth, width=\columnwidth]{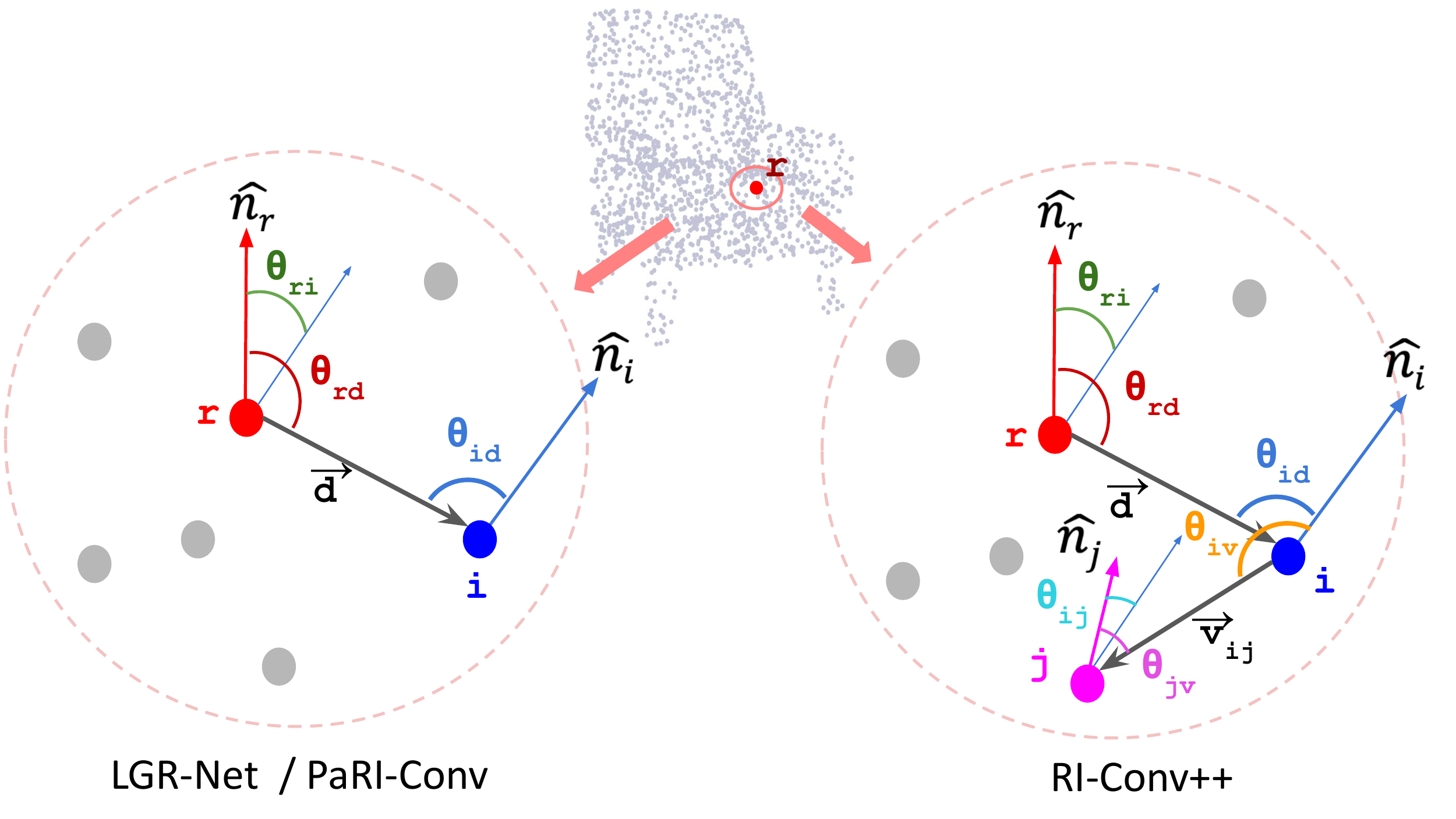}
    \caption{Figure shows the difference between shape encodings employed by the three rotation-invariant methods: LGR-Net, PaRI-Conv and RI-Conv++. A local patch around a reference point \textcolor{red}{$r$} is obtained using k-NN search, shown by the red circles. All angles $\theta$ follow notation: $\theta_{xy}$: the angle between normal ($\hat{n}$) at point $x$ and the vector $\vec{y}$. 
    LGR-Net and PaRI-Conv (shown on the left) encode relationships between a pair of points i.e. a reference point \textcolor{red}{$r$} and a neighborhood point \textcolor{blue}{$i$}, whereas RI-Conv++ (shown on the right) encodes relationship between a reference point \textcolor{red}{$r$} and a neighborhood point \textcolor{blue}{$i$}, as well as \textcolor{blue}{$i$} and its \emph{adjacent} point \textcolor{magenta}{$j$}. Such triplet-based encoding allows RI-Conv++ to learn better features towards achieving the property of rotation invariance, outperforming the other two pair-wise encoding methods, in almost all experimental settings (see Table \ref{table:final_half_so3 rotation}).
    }
    \label{fig:archi_diff}
\end{figure}

\mypara{Effect of easy/medium/hard negative pairs.}
As discussed in previous paragraphs, SO(3) rotations are the most difficult to model for instance classification task. We go a step further and analyze the effect of positive pairs, and easy, medium and hard negative pairs in the SO(3) rotation space. This setting is coupled with 50\% cut objects. \Cref{table:final_half_so3 rotation} shows quantitative results for this experiment. 
From the table, we see that as the difficulty level of negative pairs increases, the performance decreases. This is expected since objects in medium and hard negative pairs are geometrically similar, making it difficult to distinguish them. 
We also notice that training a one-layer neural network on pre-trained features boosts the performance compared to using thresholding. For all three levels of negative pairs, we notice that RI-Conv++ outperforms all other methods, with PaRI-Conv being the closest competitor due to reasons explained in Figure \ref{fig:archi_diff}.

\Cref{fig:qualitative_results} shows a few visualizations of positive and very hard negative pairs. The first and third columns show results on positive pairs, while the second and fourth columns show results on negative pairs. 
The negative pair in the second column is a pair of \textit{tables} with similar geometries but with different lengths and positions of legs (one has legs near the rim, and the other has them slightly inwards). The negative pair in the last column is a pair of \textit{ceiling lamps} with different fine-grained details. Predictions reported in \Cref{fig:qualitative_results} are calculated using \textit{Thresholding}. 
We observe that for complete objects, the frequency of correct classification by rotation-invariant methods is high even though the pairs are geometrically alike. However, they fail to distinguish positive pairs from negative ones when the input object is partially cut, revealing that these deep learning-based methods may very well be invariant to rotations but are bottlenecked by the degree of input object observations. 

In summary, the results show that the 3D Zernike, a traditional rotation invariant method, performs better than deep learning-based \emph{rotation-sensitive} methods. However, it is inferior to deep learning-based \emph{rotation-invariant} methods.

\mypara{Why does RI-Conv++ perform better in most cases?}
Deep rotation-invariant methods, LGR-Net, RI-Conv++ and PaRI-Conv, all use point coordinates with normals as input and are trained for 3D shape classification tasks. 
These methods all encode local neighborhood information for every point in the point cloud using different attributes, with some key differences, as shown in \Cref{fig:archi_diff}. Specifically, LGR-Net and PaRI-Conv consider relative angles and distances between a pair of points i.e. a reference point $r$ and its neighboring point $i$, whereas RI-Conv++ \emph{additionally} incorporates a relationship between $i$ and its \emph{adjacent} point $j$. In other words, RI-Conv++ considers a triplet of points, \{$r$, $i$, $j$\}, for feature encoding vs. \{$r$, $i$\} in LGR-Net and PaRI-Conv. This extra information about the local neighborhood makes the features learned by RI-Conv++ more informative and leads to imporved performance. This is an important observation that can help us make informed decisions when designing future rotation-invariant network architectures.
\section{Conclusion}
We analyse different approaches to the problem of identifying if two 3D objects are the rotated instances of the \emph{same} object.
%
%
We explore eight different 3D shape analysis methods, which include a traditional method, four prominent deep learning-based \emph{rotation-sensitive} methods, and three recently proposed deep learning-based \emph{rotation-invariant} methods. We evaluate these shape encoding methods in forty-eight experimental settings based on different degrees of rotational transformations, percentage of partial observations, and level of difficulty of negative pairs. 

Our study underscores the challenge in developing shape descriptors that are rotation invariant in the SO(3) space. The results, \q{on a synthetic dataset}, show that rotation-invariant 3D shape analysis methods are effective in easy cases with easy-to-distinguish input pairs. However, their performance is significantly affected when the difference in rotations on the input pair is large, or when the percentage of observation of input objects is reduced, or the difficulty level of the input pair is increased. We notice that rotation-invariant methods outperform all other methods, including the traditional rotation-invariant method (3D Zernike descriptors). And, this traditional method outperforms deep-learning-based rotation-sensitive methods. 
Finally, we provide insights by explaining the key difference in these encodings that enables RI-Conv++ to outperform its closest competitor, PaRI-Conv, on most of the experimental settings.  With this work, we hope to inspire research in this direction, and provide useful pointers in making design choices for modeling rotation-invariance for shape encoding and related downstream applications.

{\small
\bibliographystyle{ieee_fullname}
\bibliography{references}
}

\end{document}